\documentclass[a4paper]{article}

\usepackage{INTERSPEECH2022}
\usepackage{hyperref}

\title{Code Switched and Code Mixed Speech Recognition for Indic languages }
\name{Harveen Singh Chadha$^1$, Priyanshi Shah$^1$, Ankur Dhuriya$^1$, Neeraj Chhimwal$^1$, Anirudh Gupta$^1$ ,  Vivek Raghavan$^2$}
\address{
	$^1$ThoughtWorks\\
	$^2$EkStep Foundation}
\email{\{harveen.chadha, priyanshi.shah,  ankur.dhuriya, neeraj.chhimwal, anirudh.gupta \}@thougtworks.com, vivek@ekstep.org}

\begin{document}
\maketitle

\begin{abstract}
Training multilingual automatic speech recognition (ASR) systems is challenging because acoustic and lexical information is typically language specific. Training multilingual system for Indic languages is even more tougher due to lack of open source datasets and results on different approaches. We compare the performance of end to end multilingual speech recognition system to the performance of monolingual models conditioned on language identification (LID). The decoding information from a multilingual model is used for language identification and then combined with monolingual models to get an improvement of $50\%$ WER across languages. We also propose a similar technique to solve the Code Switched problem and achieve a WER of $21.77$ and  $28.27$ over Hindi-English and Bengali-English respectively. Our work talks on how transformer based ASR especially wav2vec 2.0 can be applied in developing multilingual ASR and code switched ASR for Indic languages. 

\end{abstract}

\section{Introduction}
Speech recognition has made remarkable progress in the past few years especially after the advent of deep learning. End to End networks become an attractive choice for multilingual ASR's and code switching ASR since they combine the acoustic model, pronunciation and lexicon model into a single network. This is of particular importance for countries such as India which has 22 official languages with an additional 1500 minor languages/dialects. 

\subsection{Multilingual ASR}
The use of multilingual ASR models which simultaneously transcribe many languages has gained a lot of research interest in the past \cite{5495646, SCHULTZ200131, 6639348}. Multilingual models can not only simplify pipelines in a production setting but also training on small set of similar languages can significantly improve recognition. A straight forward approach for dealing with many languages at the same time would be to identify the language with the help of a language identification system and choose the appropriate monolingual model for decoding \cite{kumar2005multilingual}. 

To train a multilingual ASR model we can take a union over all the language characters and jointly train a model on all the languages. Our approach is based on the wav2vec $2.0$ \cite{baevski2020wav2vec}.  We solve the problem of multilingual ASR system for six Indic languages: Hindi, Marathi, Odia, Tamil, Telugu, Gujarati. We train both, a multilingual ASR model for all 6 languages combined, and also monolingual models for each language. To create a language identification model(LID) we analyze the transcripts of multilingual ASR model and find out that including a LID before monolingual models gives an improvement of $51$\% for the final system.

\subsection{Code Switching ASR}
Bilingual speakers tend to have the ability to switch linguistically according to situational changes. When a bilingual speaker articulates  two  languages  successively  in  the  same discourse, there is interference between the two modes of speaking. When  two  languages  are  articulated  in  this manner, they differ significantly from the same languages as spoken in their separate social systems. This kind of interference  on  the  discourse  level  is  known  as code-switching\cite{dey2014hindi}.

Code-Switching is more commonly used in a non-literary way. Hence there is a very limited amount of code-switching corpus\cite{nj2020investigation}. Several challenges appear in this area, including the lack of training data for language modeling\cite{adel2015syntactic}, the co-articulation effects\cite{vu2012first}, and the need of expert linguistic knowledge. We solve code switching problem between two Indic language pairs: Hindi-English and Bengali-English. We use the wav2vec 2.0 for modelling this problem as well.

\section{MUCS Dataset}

This dataset\footnote{\url{https://navana-tech.github.io/MUCS2021/data.html}} was a part of MUCS 2021\cite{Diwan_2021} competition held at Interspeech 2021. The competition organizers in addition to the data provided baseline models as well. For multilingual, total duration of train set is 403 hours with sentence-wise file transcriptions. Every transcript contains text in a single script. For code switching, total duration for training set is 136 hours and every transcript contains text in two scripts pairs.

\begin{table}[!htp]\centering
	\scriptsize
	\caption{MUCS Dataset Description}\label{tab1: }
	\begin{tabular}{lrrrrr}\toprule
		Task &Language &\multicolumn{3}{c}{Duration (hrs)}  \\\cmidrule{3-5}
		& &Train &Dev &Blind \\\midrule
% 		\multirow{6}{*}{Multilingual}
        Multilingual
		&Hindi &95.05 &5.55 &5.49 \\
		&Marathi &93.89 &5.59  &0.67\\
		&Odia &94.54 &5.49  &4.66\\
		&Tamil &40 &5 &4.41 \\
		&Telugu &40 &5  &4.39\\
		&Gujarati &40 &5  &5.26\\\midrule
% 		\multirow{2}{*}{Code Switching}
        Code Switching
		&Hindi-English &89.85 &5.18 &6.24\\
		&Bengali-English &46.11 &7.02  &5.53\\
		\bottomrule
	\end{tabular}
	
\end{table}

\subsection{Data Preprocessing}
\subsubsection{Audio Data Preprocessing}
For Hindi, Marathi and Odia, we upsample the audios to 16kHz from 8kHz. Although this is not going to add any information, this is done to keep a standard sample rate across different languages. Also, the ASR algorithm we use works on 16kHz audio data. Additionally, we also perform loudness normalization of all audios in the training set.

\subsubsection{Text Data Preprocessing}
We clean all the audios for any punctuations and special characters which have no spoken form. Apart from this we also clean symbols which occur rarely in the training data. Any symbol having less than 10 frequency was removed from the training transcripts and hence ASR vocabulary too.

\subsubsection{Language Merger}
After analyzing the vocabulary of unique characters in different languages, one interesting finding was that there is just a difference of 1 character between Hindi and Marathi. So we decide to merge both the languages together during the multilingual/monolingual trainings. We denote this merger by HindMara in the sections below. Table 2 denotes the final vocab length.

\begin{table}[!htp]\centering
	\scriptsize
	\caption{Vocab length Description}\label{tab1.1: }
	\begin{tabular}{l|rrrrr}\toprule
		Lang &HindMara &Odia &Tamil &Telugu &Gujarati \\\midrule
		Vocab length &71 &64 &48 &63 &65 \\
		\bottomrule
	\end{tabular}

\end{table}

\section{Methodology}
We use wav2vec 2.0 \cite{baevski2020wav2vec} for modelling both our setups. We built our own experimentation platform\footnote{\url{https://github.com/Open-Speech-EkStep/vakyansh-wav2vec2-experimentation}} on top of fairseq toolkit for the entire modelling process. wav2vec 2.0 has 2 stages: pretraining and finetuning. In pretraining, the speech input is masked in the latent space and a contrastive task with predictions from the transformer and quantized latent speech representations\cite{baevski2020wav2vec} is solved to learn contextualized information. This enables learning powerful representations from speech audio alone. Pretraining is then followed by finetuning using the labelled data. The general methodology across experiments in both the problems remains same.

\subsection{Pretraining}
We experiment with two pretrained models. One is \textbf{Hindi-4200}, trained on 4200 hours of unlabelled Hindi data\cite{clsril}\footnote{\url{$https://github.com/Open-Speech-EkStep/vakyansh-models\#pretrained-asr-models$}} and is based on base architecture which has $95 M$ parameters.  Another model that we experiment is \textbf{XLSR-53}\cite{xlsr} which is trained on 56k hours of data across 53 languages and is based on large architecture which has $310 M$ parameters.

\subsection{Finetuning}
Pretrained models are fine-tuned by adding a fully connected layer on top of the context network with the size of output layer equal to the vocabulary of the task. Models are optimized using a CTC loss \cite{Graves06connectionisttemporal}. We finetune on both the pretrained models as described in 3.1 and perform unilingual and multilingual trainings using Vakyansh \cite{chadha2022vakyansh} toolkit.

\subsection{Language Model}
The output from speech recognition model is fed to a statistical KenLM\cite{heafield2011kenlm} based language model where a beam search is performed to find the appropriate words and correct spellings of some common words. We use IndicCorp to train most of our 5-gram KenLM based language models. The corpus for LM training is cleaned of all the characters/words that do not appear in the vocabulary of ASR training. The words are sorted based on their frequency and the top 5,00,000 words are picked. The probabilities are then calculated using 1-gram, 2-gram upto 5-gram.

\section{Problem Formulation}
\subsection{Multilingual ASR}
\subsubsection{Approach M1: Train one common multilingual model}
This is the most straightforward approach, lets say we have $N$ languages $(L_1, L_2,...,L_N)$. Each language $L_i$ has $C_i$ characters. We train our common multilingual model on combined character set $\cup_{i=1}^{N} C_i$ and a union of all training samples in fine-tuning. The model is trained by sharing parameters across all languages. We finetune on both the Hindi-4200 and XLSR-53 pretrained models. We also train a KenLM based $5$ gram language model (LM) by combining text from all the languages. 

\subsubsection{Approach M2: Language Identification + Monolingual Models}

The key part here is to utilize the common multilingual model trained in 4.1.1 as a language identification model. When a single model is capable of understanding different languages, it means it is capable of identifying the language first and then performing the transcript decoding. If we can extract the language part somehow from this model then this could be further used in the pipeline to select a monolingual model to run on. We propose the following rule based approach for Language Identification.

\begin{enumerate}
  \item Decode the audio with common multilingual model and get transcripts.
  \item Classify every character in the transcript into a language based on appearance of that character in a particular language's vocab. Since the vocabs are non overlapping we get exact counts of different languages present in the transcript. Then select the language that appears the most, this is a simple majority-voting classification technique.
  \item Once the audios are classified by their language labels predicted in the previous step, we infer them using monolingual models trained for the particular language. The idea behind using monolingual model here is that it will be more accurate than a multilingual model.
  
\end{enumerate}
There is one problem though, if the language of the audio is identified incorrectly, then we will never be able to get the correct transcript for that audio.

\subsection{Code Switching}
\subsubsection{Approach C1: One common model}
The main idea behind this approach is to include more data for English as it is the common language between different pairs. The final model will be capable to output prediction in Hindi-English or Bengali-English pairs. We finetune on both Hindi-4200 and XLSR-53 followed by a 5-gram combined language model.

\subsubsection{Approach C2: Individual models on different pairs}
The idea behind this experiment was to find out if monolingual model trained on a pair will outperform a multilingual model trained on combination of different pairs with one common language between the pairs.

% \subsection{Model Architecture}
% \subsection{Training Setup and Evaluation}
\section{Results}
\subsection{Multilingual ASR}
\subsubsection{Approach M1}
Results for Approach M1 for different pretrained models is shown in Table 3. To evaluate the models, we report the WER (Word Error Rate). The takeaway from here was that a bigger finetuned model based on XLSR-53 was performing better on the blind set. So we decide to use XLSR-53 as the common Multilingual model for Language Identification task. The performance on Marathi seems to be poor but actually the poor results were due to presence of blindset in a different format.

\begin{table}[!htp]
    \centering
    \scriptsize
    \caption{Approach M1 WER results}
    \begin{tabular}{l|r|r|r|r} \toprule
        Language &\multicolumn{2}{c}{Hindi-4200} &\multicolumn{2}{c}{XLSR-53} \\ \cmidrule{2-5}  
        ~ &Dev &Blind &Dev &Blind \\ \midrule
        Hindi &\textbf{21.51} &\textbf{18.14} & 24.34 & 25.02 \\ 
        Marathi &\textbf{15.66} & 103.68 & 16.02 &\textbf{62.03} \\ 
        Odia & 33.79 &\textbf{29.44} &\textbf{31.71} & 32.66 \\ 
        Tamil &\textbf{28.61} & 63.53 & 29.72 &\textbf{35.17} \\ 
        Telugu &\textbf{26.39} & 41.33 & 28.7 &\textbf{31.69} \\ 
        Gujarati &\textbf{15.79} & 74.87 & 17.2 &\textbf{32.5} \\ \midrule
        Average &\textbf{23.62} & 55.16 & 24.61 &\textbf{36.51} \\ 
        \bottomrule
    \end{tabular}
    
\end{table}

\subsubsection{Approach M2}
For approach M2 to work we need three things: a common multilingual model, language identification majority voting classifier, monolingual models. Common Multilingual model as explained in previous step is XLSR-53 and language identification is performed on the output transcripts from XLSR-53. 

For monolingual models we train 5 models as Hindi and Marathi were merged to form HindMara. The pretrained model here was Hindi-4200 as it is more suited for less data. During this training, we also try Data augmentation and we see $17\%$ improvement in WER for Tamil and HindMara but not so for other languages. Augmentations like volume, pitch, pace and Gaussian noise were done twice on each audio sample to get 3x training data. For language models, we use IndicCorp to create 5-gram pruned statistical models. The beam width was set to 3000 during the decoding part so as to get the best possible output with lower WER.

\begin{table}[!htp]
    \centering
    \scriptsize
    \caption{Approach M2: Monolingual model results on dev set}
    \begin{tabular}{lrrrr} \toprule
        Language &Augmentation &\multicolumn{2}{c}{WER} \\ \cmidrule{3-4}  
        & &Without LM &With LM  \\ \midrule
        HindMara	&Yes	&16.08	&\textbf{14.07} \\ 
        Odia	&No	&\textbf{27.17}	&29.1 \\ 
        Tamil	&Yes	&30.84 &\textbf{21.43} \\ 
        Telugu	&No	&29.39	&\textbf{20.63} \\ 
        Gujarati	&No	&22.53	&\textbf{20.53} \\ \midrule 
        Average	 &	&25.202	&\textbf{21.152} \\
        \bottomrule
    \end{tabular}
    
\end{table}

The key takeaway from here is that we can have different configurations for monolingual models based on what works well on which language. From the results, we decide not to use a LM with Odia as it increased the WER. Using this setup as the final setup we get an average WER of 26.56 which is an improvement of $51$\% over Approach 1. Final Results are in table 5.

\begin{table}[!htp]\centering
	\scriptsize
	\caption{Baseline and Blind Set WER of final system}\label{tab5: }
	\begin{tabular}{lrrrr}\toprule
		Language &Baseline & Our Results \\
		\midrule
		Hindi &37.2 &\textbf{12.24} \\
		Marathi &\textbf{29.04} &39.74 \\
		Odia &38.46 &\textbf{27.10} \\
		Tamil &34.09 &\textbf{27.20} \\
		Telugu & 31.44 &\textbf{22.43}\\
		Gujarati &\textbf{26.15} &30.65 \\
		
		\midrule
		Average &32.73 &26.56 \\
		\bottomrule
	\end{tabular}
	
\end{table}

\subsection{Code Switching ASR}
In code switching dataset, there were multiple instances of the same English word appearing both in the Latin script and the native scripts of Hindi and Bengali in the training data. Given this inconsistency in script usage for English words, the ASR predictions of English words could either be in the native script or in the Latin script. To allow for both English words and their
transliterations in the respective native scripts to be counted as correct during the final word error rate computations, we calculated transliterated WER (T-WER) metric along with the standard WER metric. T-WER will count an English word in the reference text as being correctly predicted if it is in English or in its transliterated form in the native script

\subsubsection{Approach C1}
Results from Approach C1 are in table 6. Training data was combined from both pairs and even for the LM training, the training text corpus was combined to create the final LM. The key takeaway from here was that the finetuned model on Hindi-4200 outperformed XLSR-53 again due to the presence of Hindi data in the pretrained model. 

\begin{table}[!htp]\centering
		\scriptsize
		\caption{Approach C1: Results on blindset}\label{tab2: }
		\begin{tabular}{lrrrr}\toprule
			Language  &\multicolumn{2}{c}{Hindi-4200} &\multicolumn{2}{c}{XLSR-53}  
			\\\cmidrule{2-5} 
			& WER &T-WER & WER &T-WER\\\midrule
			Hindi-English 23.11 &21.76 &24.77 &22.67\\
			Bengali-English  &29.83 &28.88 &31.83  &30.81\\
			\midrule
			Average &26.47 & 25.32 &28.3 &26.74\\
			\bottomrule
		\end{tabular}

\end{table}

\subsubsection{Approach C2}
Based on the findings from previous Approach, the idea behind this approach was to train separately on Hindi-English pairs and Bengali-English pairs. The main difference between Multilingual ASR and code switched is that during the blindset evaluation, language for multilingual ASR is not provided (it is getting calculated automatically) which is expected from it as well given the name but during evaluation of code switched ASR the pair language is available explicitly.

\begin{table}[!htp]\centering
	
		\scriptsize
		\caption{Approach C2: Results using Hindi-4200 on blindset}\label{ta32: }
		\begin{tabular}{lrrrr}\toprule
			Language &\multicolumn{2}{c}{Baseline} &\multicolumn{2}{c}{Our Results} 
			\\\cmidrule{2-5} 
			& WER &T-WER & WER &T-WER  \\\midrule
			Hindi-English & 25.53 &23.80 & 21.77 & 20.75\\
			Bengali-English &32.81 &31.70 & 28.27 & 26.96 \\
			\midrule
			Average &29.17 &27.75 &25.02 & 23.85 \\
			\bottomrule
		\end{tabular}
	
\end{table}

\section{Conclusion and Future Work}
For the multilingual ASR, through our solution we were able to beat the baseline provided by competition organizers easily as shown in Table 5. In two languages Marathi and Gujarati, we were not able to get better results. For marathi, later it was on clarified by the competition organizers that the collection format of data was wrong, hence it was excluded from the final rankings.

It has been usually the case with multilingual ASR systems that combined training on multiple languages is able to outperform individual end to end systems for each language. Using pre-trained model in a high resource language together with LID we are able to show that low resource languages can benefit greatly from a high resource language. In the recent times code switching problems got better results when there was a frame level language identification information used as a condition for transcribing the final output.

\section{Acknowledgements}

All authors gratefully acknowledge Ekstep Foundation for supporting this project financially and providing infrastructure. A special thanks to Dr. Vivek Raghavan for constant support, guidance and fruitful discussions. We also thank Rishabh Gaur, Ankit Katiyar, Anshul Gautam, Nikita Tiwari, Heera Ballabh, Niresh Kumar R, Sreejith V, Soujyo Sen and Amulya Ahuja for automated data pipelines and infrastructure support for model training and model testing. 

\bibliographystyle{IEEEtran}

\bibliography{code_switched}

% \begin{thebibliography}{9}
% \bibitem[1]{Davis80-COP}
%   S.\ B.\ Davis and P.\ Mermelstein,
%   ``Comparison of parametric representation for monosyllabic word recognition in continuously spoken sentences,''
%   \textit{IEEE Transactions on Acoustics, Speech and Signal Processing}, vol.~28, no.~4, pp.~357--366, 1980.
% \bibitem[2]{Rabiner89-ATO}
%   L.\ R.\ Rabiner,
%   ``A tutorial on hidden Markov models and selected applications in speech recognition,''
%   \textit{Proceedings of the IEEE}, vol.~77, no.~2, pp.~257-286, 1989.
% \bibitem[3]{Hastie09-TEO}
%   T.\ Hastie, R.\ Tibshirani, and J.\ Friedman,
%   \textit{The Elements of Statistical Learning -- Data Mining, Inference, and Prediction}.
%   New York: Springer, 2009.
% \bibitem[4]{YourName17-XXX}
%   F.\ Lastname1, F.\ Lastname2, and F.\ Lastname3,
%   ``Title of your INTERSPEECH 2021 publication,''
%   in \textit{Interspeech 2021 -- 20\textsuperscript{th} Annual Conference of the International Speech Communication Association, September 15-19, Graz, Austria, Proceedings, Proceedings}, 2020, pp.~100--104.
% \end{thebibliography}

\end{document}